%% file: main.tex
\documentclass[runningheads]{llncs}
\usepackage{graphicx}
\usepackage{amsmath}
\usepackage{amsfonts}
\usepackage{amssymb}
\usepackage{mathtools}
\usepackage{multicol}
\usepackage{multirow}
\usepackage{xcolor}
\usepackage{subcaption}
\usepackage{placeins}
\usepackage{enumerate}
\usepackage{hyperref}
\usepackage{bbold}

\usepackage{algorithm}
\usepackage{algpseudocode}

\usepackage{bm}

\include{defs}
\DeclareUnicodeCharacter{2212}{-}
\begin{document}
%
\title{Online 3D reconstruction and dense tracking \\
in endoscopic videos}



%
%
%
%
%


\author{Michel Hayoz\inst{1} \and Christopher Hahne\inst{1} \and Thomas Kurmann\inst{3} \and Max Allan\inst{3} \and Guido Beldi\inst{2} \and Daniel Candinas\inst{2} \and Pablo Márquez-Neila\inst{1} \and Raphael Sznitman\inst{1}}
\authorrunning{M. Hayoz et al.}

\institute{ARTORG Center, University of Bern, Switzerland \and Dept. of Visceral Surgery and Medicine, Inselspital, Switzerland \and Applied Research, Intuitive Surgical, USA \\ \email{michel.hayoz@unibe.ch}}

\maketitle              
\begin{abstract}
\input{abstract}

\end{abstract}
\input{1_intro}
\input{2_method}
\input{3_experiments}
\input{4_conclusion}
\begin{credits}
\subsubsection{\discintname}
The authors have no competing interests to declare that are relevant to the content of this article.
\end{credits}
\bibliographystyle{splncs04}
\bibliography{innosuisse}

\end{document}

%% file: defs.tex
\newif\ifdraft
\drafttrue

\definecolor{orange}{rgb}{1,0.5,0}
\definecolor{pink}{rgb}{0.98, 0.38, 0.5}
\definecolor{darkgreen}{rgb}{0.055, 0.490, 0.016} 

\definecolor{senioryellow}{rgb}{0.86, 0.8, 0.19}
\definecolor{assistantblue}{rgb}{0, 0.133, 0.31}

\ifdraft
 \usepackage[normalem]{ulem}
 \newcommand{\RS}[1]{{\color{red}{\bf RS: #1}}}
 
 \newcommand{\PMN}[1]{{\color{orange}{\bf PMN: #1}}}

\else
 \newcommand{\sout}[1]{}
 \newcommand{\RS}[1]{{\color{red}{}}}
 
 \newcommand{\PMN}[1]{{\color{red}{}}}
 
\fi

\newcommand{\real}{\mathbb{R}}
\newcommand{\quat}{\mathbb{H}}

\newcommand{\image}{\mathbf{i}}
\newcommand{\depth}{\mathbf{d}}
\newcommand{\opacity}{\mathbf{o}}
\newcommand{\x}{\mathbf{x}}
\newcommand{\f}{\mathbf{f}}
\newcommand{\p}{\mathbf{p}}
\newcommand{\q}{\mathbf{q}}
\newcommand{\bu}{\mathbf{u}}

\renewcommand{\d}{\mathbf{d}}

\newcommand{\G}{\mathcal{G}}

\newcommand{\bP}{\mathbf{P}}
\newcommand{\calK}{\mathcal{K}}

\newcommand{\bmu}{\bm{\mu}}

\newcommand{\comment}[1]{}

\usepackage{pifont}
\newcommand{\cmark}{\ding{51}}%
\newcommand{\xmark}{\ding{55}}%

\DeclareMathOperator*{\argmin}{argmin}
\DeclareMathOperator*{\sigmoid}{sigm}
\DeclareMathOperator*{\diag}{diag}
\newcommand{\Mat}{R}

%% file: abstract.tex
3D scene reconstruction from stereo endoscopic video data is crucial for advancing surgical interventions. In this work, we present an online framework for online, dense 3D scene reconstruction and tracking, aimed at enhancing surgical scene understanding and assisting interventions. Our method dynamically extends a canonical scene representation using Gaussian splatting, while modeling tissue deformations through a sparse set of control points. We introduce an efficient online fitting algorithm that optimizes the scene parameters, enabling consistent tracking and accurate reconstruction. Through experiments on the StereoMIS dataset, we demonstrate the effectiveness of our approach, outperforming state-of-the-art tracking methods and achieving comparable performance to offline reconstruction techniques. Our work enables various downstream applications thus contributing to advancing the capabilities of surgical assistance systems. 

%% file: 1_intro.tex
\section{Introduction}
\label{sec:intro}
3D scene reconstruction represents a fundamental challenge for surgical scene understanding~\cite{bodenstedt2018comparative,Kurmann21,hayoz2023pose,MaierHein2022}. The ability to infer accurate 3D geometry from endoscopic image data would have numerous important downstream tasks, such as retrospective analysis for surgical training, integrated virtual overlays of pre-operative image data, and augmented surgical robotics. As such, the need for methods that yield real-time and consistent 3D estimates of the surgical site is paramount for the next generation of surgical assistant tools.

Recent years have seen a variety of highly promising methods for 3D reconstruction of surgical scenes. Primarily driven by advances in neural rendering, these have shown an impressive ability to reconstruct dynamically deforming surgical scenes~\cite{endonerf,endosurf,NeuraLerPlane,OrthogonalPlane,SemanticSuper}. Unfortunately, they are limited by offline processing or lack physical constraints to control tissue deformations. Other approaches originating from natural images, such as video-based tracking~\cite{NEURIPS2022_58168e8a,zheng2023point} and multi-camera reconstruction~\cite{luiten2023dynamic}, have also shown great potential, while their refinement to surgical scenes is not always evident. The recently introduced benchmark for soft-tissue trackers in robotic surgery~\cite{CARTUCHO2024102985} will also provide interesting algorithmic developments despite its focus on sparse-point tracking and not dense tracking -- the latter being essential for most downstream applications.

In this work, however, we focus on online 3D reconstructions from stereo endoscopic video data. Drawing from recent developments in Gaussian splatting~\cite{kerbl3Dgaussians}, which allows for fast reconstruction and rendering, we propose a novel framework by way of dense point tracking in the endoscopic scene. Unlike traditional methods that assume a fixed topology at initialization~\cite{luiten2023dynamic,endonerf}, our approach dynamically initializes a set of Gaussian models and updates as new scene parts become visible over time. We incorporate per-Gaussian learning rate modulation to ensure accurate optimization while retaining information from past frames. Additionally, we integrate optical flow motion initialization to ease convergence in single-camera settings. Finally, we parametrize deformations using a small set of control points distributed on the scene surface and proportional to scene complexity. The reduced number of control points, allowed by using Gaussian kernel interpolation in the deformation fields, results in faster fitting times, simpler geometric priors, and quicker tracking. To validate our method, we evaluate it on the publicly available StereoMIS dataset. We outperform state-of-the-art tracking algorithms and demonstrate comparable performance to offline 3D reconstruction methods~\footnote {Code: \url{https://github.com/mhayoz/online_endo_track}, data:\url{https://zenodo.org/records/10867949}}.

%% file: 2_method.tex
\section{Method}

\begin{figure}[!t]
\centering
\includegraphics[width=\textwidth]{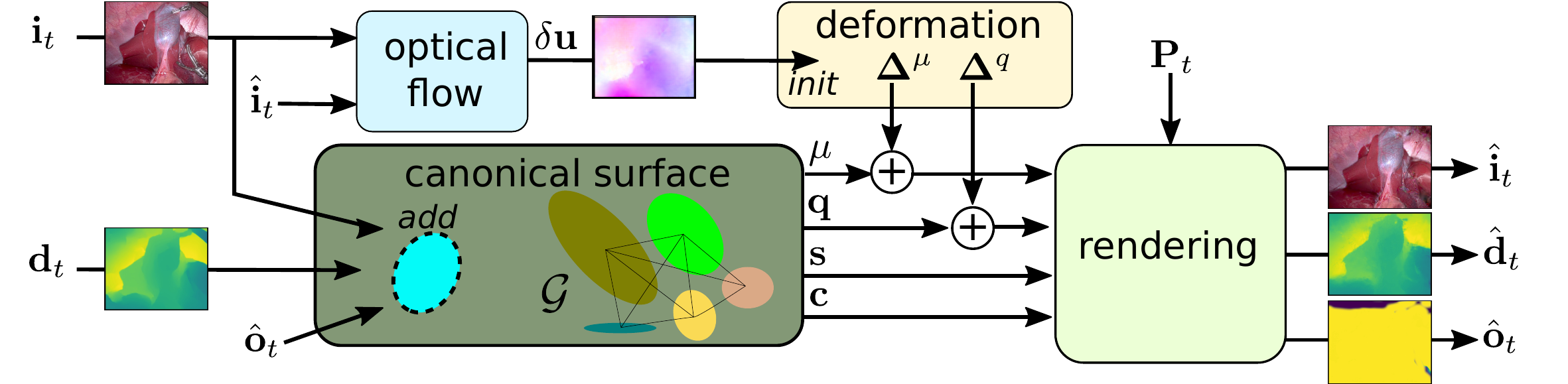}
\caption{Overview of our proposed scene reconstruction and dense tracking method.}
\label{fig1}
\end{figure}

Our scene reconstruction framework is designed to densely track surface points in a video sequence (Fig.~\ref{fig1}). Each frame~$t$ of the video consists of tuple~$\f_t = (\image_t, \depth_t, \bP_t)$ containing the RGB image, the depth map, and the camera pose, respectively. Our tracking method models the scene using a combination of static Gaussian splatting and non-rigid deformations. The parameters of the model are optimized in an online manner based on photometric, geometric, and physical constraints by minimizing the reconstruction error between each video frame~$\f_t$ and the synthetically generated frame~$\hat{\f}_t$. The subsequent sections describe the scene representation model and the online fitting algorithm. 

\subsection{Scene Model}
\subsubsection{Canonical and non-rigid scenes.} The rigid component of the scene, known as \emph{canonical scene}, is modeled via Gaussian splatting~\cite{kerbl3Dgaussians} using a collection~$\G=\{g_i\}_{i=1}^{G}$ of 3D colored Gaussians. A colored Gaussian is defined by a tuple~$g_i$ containing the position~$\bm{\mu}_i\in\real^3$, the scale~$\mathbf{s}_i\in\real^3$, the orientation~$\mathbf{q}_i\in\quat$, and the color~$\mathbf{c}_i\in\real^3$ of the Gaussian. The covariance of the Gaussian can be trivially computed from its orientation and scale as
    $\mathbf{\Sigma} = \Mat(\mathbf{q}) \diag(\mathbf{s})^2 \Mat(\mathbf{q})^T$,
where $\Mat(\mathbf{q})$~is the rotation matrix corresponding to the quaternion~$\mathbf{q}$.

Tissue deformations are an integral part of surgical scenes. We model these by warping the canonical surface with a translation vector field~$\bm{\Delta}^\mu\colon \real^3\to \real^3$ and a rotation vector field~$\bm{\Delta}^q\colon \real^3\to \quat$. 
These warps act additively over the locations and orientations of the Gaussians of the canonical scene. For the Gaussian~$g_i$, its warped location and orientation are then~$\bm{\mu}'_{i}=\bm{\mu}_i + \bm{\Delta}^\mu(\bm{\mu}_i)$, and $\mathbf{q}'_{i}=\mathbf{q}_i+\bm{\Delta}^q(\bm{\mu}_i)$, respectively. Note that the scales of the Gaussians are not warped, as we expect nearly isometric tissue deformation.

The deformation fields are modelled with a collection~$\calK=\{(\p_k, \delta\bm{\mu}_k, \delta\mathbf{q}_k)\}_{k=1}^{K}$ of control points. Each control point is a tuple containing its position~$\p_k\in\real^3$, a translation offset~$\delta\bm{\mu}_k\in\real^3$, and an orientation offset~$\delta\mathbf{q}_k\in\quat$, which serve as the parameters of the translation and the orientation fields. Similar to~\cite{dynamicfusion}, the translation field at the location~$\x\in\real^3$ is defined as a weighted average,
\begin{equation}
\bm{\Delta}^\mu(\x)= \frac{1}{\sum_{k=1}^{K} w(\x, \p_k)}\sum_{k=1}^{K} w(\x, \p_k) \delta\bm{\mu}_k,
\end{equation}
where the Gaussian kernel,
$w(\x_1, \x_2) = \exp\left(-\gamma \left\| \x_1 - \x_2 \right\|_2^2\right)$,
is used to measure the contribution of control point~$k$ to location~$\x$, and the hyperparameter~$\gamma$ determines the rate of decay in the influence of the control points over distance. The orientation field is defined similarly using the orientation offsets of the control points.

\subsubsection{Rendering.} Following~\cite{kerbl3Dgaussians}, the image rendering function~$r_{\textrm{image}}$ takes a collection of Gaussians~$\G$ and the camera pose~$\bP$ and computes the color of the pixel at location~$\bu\in\real^2$ as,
\begin{equation}
    r_{\textrm{image}}(\bu; \G, \bP) = \sum_{i=1}^{G} \mathbf{c}_{\omega[i]}\cdot \alpha_{\omega[i]} \prod_{i'=1}^{i-1} \left(1 - \alpha_{\omega[i']}\right),
\end{equation}
where sequence~$\omega$~contains the indices of the Gaussians sorted by depth from the camera. The factor~$\alpha_i$ is the opacity corresponding to Gaussian~$g_i$ at pixel~$\bu$ after projection onto the screen space. Details on computing~$\omega$ and Gaussian projection can be found in~\cite{kerbl3Dgaussians}. Analog to image rendering, the depth rendering function replaces color with depth, while the opacity rendering function replaces color with a constant value of 1.
We obtain the synthetic color image~$\hat{\image}(\G, \bP)$, depth~$\hat{\depth}(\G, \bP)$ and opacity~$\hat{\opacity}(\G, \bP)$ by evaluating the corresponding rendering functions at all image pixels~$\{\bu_p\}_{p=1}^P$. 

\subsection{Model fitting}

Model fitting works in an online manner. For each new frame at time~$t$, we find the model parameters~$\G_t$ and~$\calK_t$ that minimize the differences between the measured images~$\image_t, \depth_t$ and their predicted counterparts~$\hat{\image}_t, \hat{\depth}_t$. Our optimization approach proceeds in three steps: first, the canonical scene is updated with new Gaussians covering previously unseen areas of the scene; second, control points~$\calK_t$ are initialized using optical flow; third, the parameters in~$\G_t$ and~$\calK_t$ are jointly optimized to minimize the reconstruction error.

{\bf Canonical scene extension:} As the camera moves throughout the sequence, establishing a canonical scene at the beginning of the sequence, as is done in~\cite{luiten2023dynamic}, is not feasible. Instead, we progressively expand the canonical scene by adding new Gaussians to cover new regions of the scene as they appear. We identify these regions by finding pixels with low opacity in the opacity image~$\hat{\opacity}(\bm{\Delta}(\G_{t-1}, \calK_{t-1}), \bP_t)$, and add a new Gaussian to each pixel with an opacity smaller than~$0.95$. The parameters of the new Gaussian are set according to the color and coordinates of the corresponding pixel~$\bu$. Its position is found by back projection to the 3D scene, $\bmu = \pi_{3D}(\bu; \depth_t, \bP_t)$, and its scale is set to the distance between~$\bmu$ and to the position~$\bmu_j$ of the nearest Gaussian.

{\bf Control point initialization:} The control points of the deformation fields are placed in the locations of a random subset of $P_t$~Gaussians serving as \emph{anchor Gaussians},
    $\p_k = \bmu_{\sigma_t[k]},\quad k\in\{1, \ldots, K_t\}$,
where $\sigma_t$~contains the indices of the \emph{anchor Gaussians} obtained as $K_t$~random elements sampled without replacement from the Gaussian index set~$\{1, \ldots, G_t\}$. The number of control points (and anchor Gaussians)~$K_t$ is set to a fraction of the number of Gaussians, $K_t = \frac{G_t}{k}$. We set~$k=64$ in all our experiments.

The translation offsets~$\delta\bmu_k$ of the control points are initialized using optical flow. RAFT~\cite{Teed2021} calculates optical flow between the synthetic image~$\hat{\image}(\G_t, \bP_t)$ and the frame image~$\image_t$, yielding screen-space offsets that are projected back to the 3D space. The control point offsets~$\delta\bmu_k$ are initialized to minimize the difference between the modeled offsets and those computed with optical flow. This optimization, detailed in the supplementary material, efficiently finds a closed-form solution via least squares.

{\bf Energy minimization:} In the last stage, the initialized parameters in~$\G_t$ and~$\calK_t$ (except positions~$\{\p_k\}$) are updated to minimize the energy function,
\begin{equation}
\argmin_{\G_t, \calK_t} E_\textrm{external} + E_\textrm{internal},
\end{equation}
that describes the quality of the model fit as a combination of an external energy and an internal energy. The external energy penalizes the deviations between the observed image and depth and their synthetic counterparts, measured with the standard MSE,
\begin{equation}
    E_\textrm{external} = \lambda_1 \left\|\image_t - \hat{\image}_t\right\|^2_2 + \lambda_2 \left\|\depth_t - \hat{\depth}_t\right\|^2_2.
    \label{eq:loss}
\end{equation}
The synthetic image~$\hat{\image}_t = \hat{\image}(\bm{\Delta}(\G_t, \calK_t), \bP_t)$ and depth~$\hat{\depth}_t = \hat{\depth}(\bm{\Delta}(\G_t, \calK_t), \bP_t)$ are rendered after warping the canonical scene with the deformation fields. The internal energy
\begin{equation}
    E_\textrm{internal} = \lambda_3 E_\textrm{rigidloc} + \lambda_4 E_\textrm{rigidrot} + \lambda_4 E_\textrm{iso} + \lambda_5 E_\textrm{visible}
\end{equation}
incorporates geometric and physical priors. Unlike previous methods, these priors are applied only to pairs of neighbor anchor Gaussians rather than all Gaussian pairs. Rigidity terms penalize changes in the relative positions and orientations of neighboring anchor Gaussians,
\begin{eqnarray}
    E_\textrm{rigidloc} & = & \dfrac{1}{4K_t} \sum_{i\in\sigma_t} \sum_{j\in\mathcal{N}^4_i} w(\bmu_{i, t}, \bmu_{j, t}) \left\|(\bmu'_{j, t-1} - \bmu'_{i, t-1}) - (\bmu'_{j, t} - \bmu'_{i, t})\right\|^2_2, \\
    E_\textrm{rigidrot} & = & \dfrac{1}{4K_t} \sum_{i\in\sigma_t} \sum_{j\in\mathcal{N}^4_i} w(\bmu_{i, t}, \bmu_{j, t}) \left\|(\q'_{j, t-1}\q_{i, t-1}^{\prime -1}) - (\q'_{j, t}\q_{i, t}^{\prime -1})\right\|^2_2,
\end{eqnarray}
where $\sigma_t$~contains the indices of the anchor Gaussians and~$\mathcal{N}^4_i$ contains the 4~nearest anchor Gaussians to~$i$. Similarly, the isometry term encourages the translation field to produce nearly isometric deformations,
\begin{equation}
E_\text{iso} = \frac{1}{4K_t}\sum_{i\in\sigma_t} \sum_{j\in\mathcal{N}^4_i} w(\bmu_{i, t}, \bmu_{j, t}) \left| \lVert  (\bmu_{j, t} - \bmu_{i, t})\rVert^2_2 - \lVert (\bmu'_{j, t} - \bmu'_{i, t})\rVert^2_2\right|.
\end{equation}
Finally, the visibility term 
\begin{equation}
E_\textrm{visible} = \frac{1}{\sum_{k=1}^{K_t} \mathbb{I}(\p_k; \bP_t)}\sum_{k=1}^{K_t} \mathbb{I}(\p_k; \bP_t) \left\|\delta\bmu_{k}\right\|^2_2
\end{equation}
penalizes deformations of the scene parts that do not project to the image, preventing drift.
The invisibility predicate $\mathbb{I}(\p; \bP)$ is~$1$ if the point $\p$ is not visible on the screen, and $0$~otherwise.

{\bf Gradient modulation:} To prevent drifting in the canonical scene, we gradually slow down the updates to the parameters of its Gaussians. To this end, we count the number of times that each Gaussian~$i$ has been updated, denoted ~$v_i$, and compute the modulation factor~$\rho_i = 2\left(1-\sigmoid (c_1 v_i - c_2)\right)$. The gradients of the energy with respect to the parameters of the Gaussian~$i$ are multiplied by this factor before applying the optimizer update rule.

{\bf Tracking:} Our method enables tracking any surface point of the scene. Given a surface point~$\x$ at time~$t$, tracking starts by approximating it with its closest Gaussian~$i$ in~$\G_t$. For subsequent frames, its 3D~position is given by~$\bmu'_i$. When the point to track is given as a screen-space 2D~point~$\bu$, we first find its corresponding surface point~$\x=\pi_{3D}(\bu; \d_t, \bP_t)$ and proceed as before.

%% file: 3_experiments.tex
\section{Experimental setup and results}

{\bf Datasets:} We evaluated our method on the StereoMIS dataset~\cite{hayoz2023pose} and selected subsequences of 200 frames with 10 frames per second with resolution 512x640 pixels. All sequences contain challenging scenes with breathing motions, tissue deformations, and occlusions. In each frame, we manually annotated 3 to 4 distinct landmarks to evaluate the tracking. We cannot benchmark our method on the SurgT-Challenge dataset~\cite{CARTUCHO2024102985} because it does not feature camera poses, an essential input to our method.

{\bf Baselines:} PIPS++~\cite{zheng2023point} is a SOTA 2D long-term tracking approach. Due to memory constraints, we partition the sequences into chunks of 50 frames. We then initialize tracked points using the last estimated locations from the preceding chunk and link all estimates to form complete trajectories. To simulate dense tracking, we estimate a uniform grid of 2048 points defined in the initial frame and linearly interpolate trajectories of the evaluation points. We mask instrument areas in the input RGB images by filling them with black.

Similar to top-performing techniques in the SurgT-Challenge~\cite{CARTUCHO2024102985}, we employ frame-to-frame optical flow to estimate dense point tracking. Specifically, we utilize RAFT~\cite{Teed2021} as a SOTA optical flow estimation method. We set the optical flow to zero for pixels occupied by surgical instruments.

{\bf Implementation Details:}
\label{sec:seg}
We optimize the canonical scene for 1000 iterations on the first frame, setting the deformation fields to zero. For each subsequent frame, we optimize for 100 iterations using Adam~\cite{adam}. We run our code on an NVIDIA RTX3090 GPU, resulting in an average processing time of 2~seconds per frame. We infer depth from stereo RGB images using the stereo disparity estimated by RAFT~\cite{Teed2021} and mask surgical instrument pixels in Eq.~\ref{eq:loss}, with masks inferred using DeepLabv3+~\cite{Chen_2018_ECCV} trained on EndoVis2018 segmentation dataset~\cite{Allan2019}. Due to the random sampling of anchor Gaussians, we run each experiment 100 times and report the average.

{\bf Metrics:}
Following~\cite{NEURIPS2022_58168e8a}, we utilize the median trajectory error (MTE), the average position accuracy~$\delta_\text{avg}$, and the survival rate, as measures for accuracy and robustness in tracking. We do not assess point tracking accuracy in 3D due to the lack of reliable ground-truth depth estimates. Similarly, we only visually assess the 3D reconstruction due to having only a single endoscope and thus no hold-out views.

\subsection{Results}
\begin{figure}[!t]
\centering
\includegraphics[width=\textwidth]{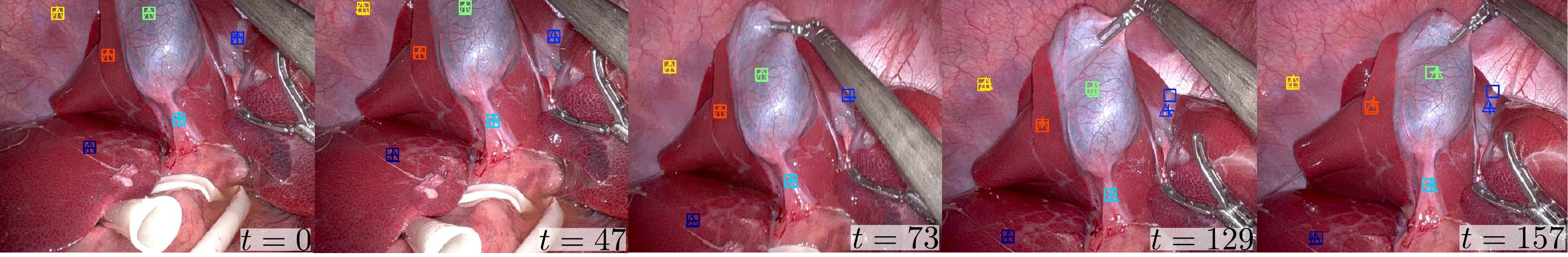}
\includegraphics[width=\textwidth]{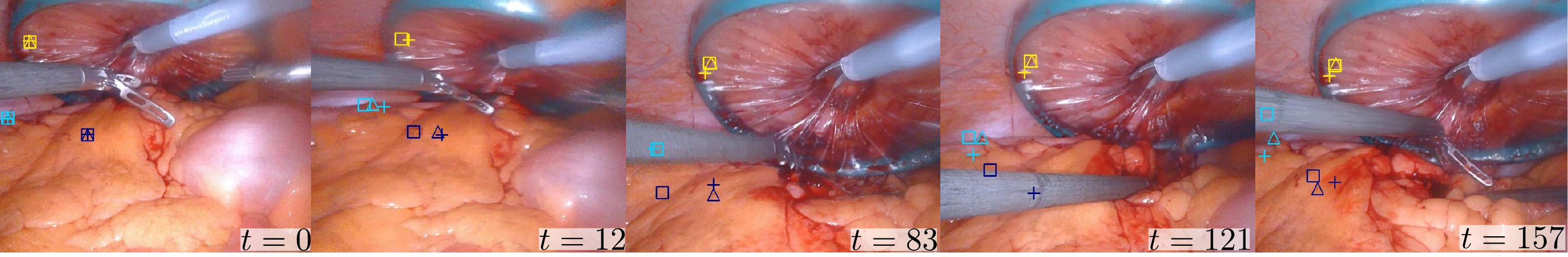}
\includegraphics[width=\textwidth]{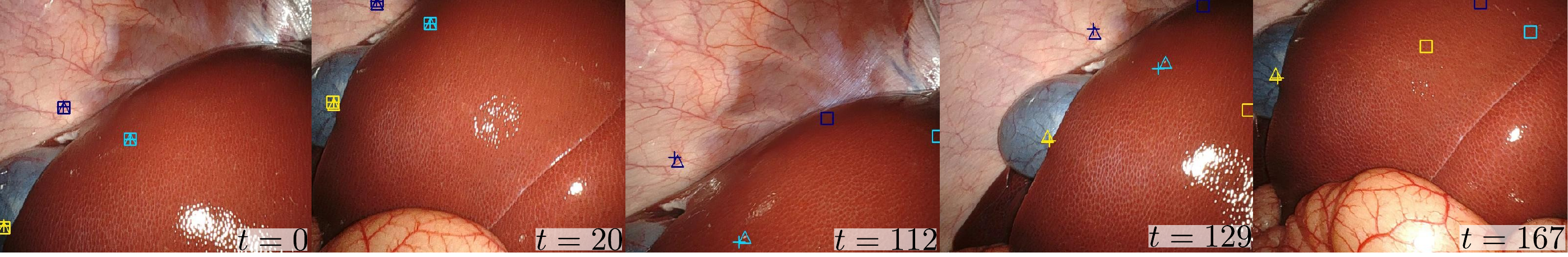}
\caption{2D point tracking over time results. Annotated ground-truth points are marked with triangles, PIPS++ with squares, and \textbf{ours} with crosses.}
\label{fig:examples}
\end{figure}

Tab.~\ref{tab:results} presents the point tracking results on the StereoMIS dataset. Our method consistently outperforms the baselines across most cases and, on average, for all three evaluation metrics. This demonstrates the accuracy captured by MTE and $\delta_{\text{avg}}$ and the robustness of our method indicated by the survival rate. Our method achieves a 100.0\% survival rate in two cases, indicating successful tracking of all points until the end of the sequences without failure. All methods accurately track points embedded on textured surfaces, as illustrated in Fig.~\ref{fig:examples}~(top). They also handle breathing motion and tool-induced deformation, which validates the physical constraints imposed by our method.
\begin{table}[t]
\begin{tabular}{cccccccccc}
Metric                          & Method              & P1\_1          & P2\_0           & P2\_1           & P3\_1           & P3\_2          & H1\_1          & H3\_1           & mean           \\ \hline
\multirow{3}{*}{MTE $\downarrow$ (px)}            & PIPS++\cite{zheng2023point}       & 67.30          & 10.40           & 317.40          & 5.50            & 21.60          & 129.20          & 10.40           & 80.26          \\
        & RAFT\cite{Teed2021}         & 42.72          & 83.86           & 197.67 & 10.98           & 18.31          & 126.18         & 21.66           & 71.63          \\
        & \textbf{ours}               & \bf{21.04} & \bf{7.91}            & \bf{14.29}           & \bf{4.14}            & \bf{14.20}          & \bf{10.51}          & \bf{8.60}   & \bf{11.53} \\\hline
\multirow{3}{*}{$\delta_\text{avg}$ $\uparrow$ (\%)} & PIPS++\cite{zheng2023point}      & \bf{42.90} & 66.20  & 31.20           & 77.80           & 81.10          & 33.90          & 66.20  & 57.04          \\
                                & RAFT\cite{Teed2021}         & 41.62          & 37.07           & 34.04           & 67.02           & 77.80          & 29.56          & 66.67           & 50.54          \\
                                
                                & \textbf{ours}              & 32.36          & \bf{66.89}           & \bf{61.14}  & \bf{80.46}           & \bf{83.67} & \bf{54.66}          & \bf{68.50}           & \bf{63.95} \\\hline
\multirow{3}{*}{survival $\uparrow$ (\%)}        & PIPS++\cite{zheng2023point}       & 37.00          & \bf{100.00}          & 57.30           & 93.80 & 82.20          & 62.40          & \textbf{100.00}         & 76.10         \\
                                & RAFT\cite{Teed2021}         & 50.33          & 53.67           & 57.33           & 89.33           & 82.25          & 59.50          & 80.63           & 67.58          \\
                                & \textbf{ours}             & \bf{70.90}& \bf{100.00} & \bf{87.45}  & \bf{100.00} & \bf{84.20}  & \bf{87.67}          & 91.13 & \bf{88.77}
\end{tabular}
\caption{Experimental results on StereoMIS dataset.}
\label{tab:results}
\end{table}

Our method handles occlusions and remains robust against motion blur caused by rapid camera movement, as illustrated in Fig.~\ref{fig:examples}~(middle) at $t=12$. In contrast, PIPS++ fails to track points with occlusions lasting longer than its chunk size, yet our method handles arbitrarily long occlusions, as depicted in Fig.~\ref{fig:examples}~(bottom), where all points are tracked even after more than 100 frame occlusions between $t=20$ and $t=112$. 

In some cases, our method struggles to capture tissue deformations accurately after long occlusions in regions with repetitive or no texture. This is attributed to unobserved tissue deformations, causing discrepancies between the estimated and actual deformations, leading to incorrect convergence, as illustrated in Fig.\ref{fig:examples}~(middle) at $t=83$ and $t=157$. Note, our method is not intrinsically limited to short sequences but tracking in a real surgical scenario may pose unaddressed challenges for long-term tracking and reconstruction of large scenes.

{\bf Comparison against offline reconstruction methods}
Tab.~\ref{tab:offline} provides an additional comparison against state-of-the-art offline endoscopic scene reconstruction methods~\cite{endonerf,endosurf}. Our method outperforms EndoNerf and achieves comparable results to EndoSurf while being online and using only a fraction of the processing time. Example images and implementation details for~\cite{endonerf,endosurf} are provided in the supplementary material.
\begin{table}[h]
\begin{tabular}{cccccc}
Method   & online & processing time & MTE $\downarrow$( px) & $\delta_\text{avg}$ $\uparrow$  (\%) & survival $\uparrow$ (\%)  \\ \hline
EndoNerf~\cite{endonerf} & \xmark & 7h &50.70$\pm$63.05 & 31.22$\pm$17.17 & 22.54$\pm$22.05 \\
EndoSurf~\cite{endosurf} & \xmark & 11h &\bf{7.78}$\pm$\bf{4.89}   & \bf{67.87}$\pm$\bf{10.41} & 87.88$\pm$14.49 \\
\bf{ours}     & \cmark & 7min & 11.53$\pm$5.51  & 63.95$\pm$17.23 & \bf{88.77}$\pm$\bf{10.00}
\end{tabular}
\caption{Comparison to offline 3D reconstruction methods on StereoMIS. Metrics are reported as the mean $\pm$ std over all sequences.}
\label{tab:offline}
\end{table}

{\bf Ablation study}
We present an ablation study in~Tab.~\ref{tab:ablation}. \textit{sparse} refers to the deformation representation using a sparse set of control points, whereas \textit{dense} explicitly represents the deformation for each Gaussian as in~\cite{luiten2023dynamic}. The most significant elements include the isometry energy~$E_{\text{iso}}$, the visible energy~$E_{\text{visible}}$, and the sparse deformation representation. While less critical, optical flow initialization and local-rigid energies still enhance effectiveness and robustness, which is evident in the increased standard deviation of metrics when omitted from optimization.
\begin{table}[]
\begin{tabular}{ccccccccc}
Description  & $E_{\text{rigid}}$ & $E_{\text{iso}}$ & $E_{\text{visible}}$ & sparse & flow & MTE $\downarrow$ (mm) & $\delta_\text{avg}$ $\uparrow$(\%)  & survival $\uparrow$ (\%) \\ \hline
\textbf{ours}                & \cmark   & \cmark&   \cmark &  \cmark&  \cmark     & \textbf{11.53}$\pm$\textbf{5.51}     & \textbf{63.95}$\pm$\textbf{17.23}     &   \textbf{88.77}$\pm$\textbf{10.00}   \\
w/o local-rigid          & \xmark   & \cmark  &  \cmark   &\cmark  &  \cmark  &  12.26$\pm$8.31  &     61.72$\pm$19.06 & 84.50$\pm$12.12 \\
w/o iso loss             & \cmark  & \xmark &  \cmark    & \cmark &  \cmark   &  17.18$\pm$9.99 & 55.68$\pm$18.17 &    78.18$\pm$18.91       \\
w/o inv. loss       & \cmark  & \cmark & \xmark      & \cmark  &  \cmark &  17.55$\pm$11.23   & 53.36$\pm19.26$     &  82.63$\pm$19.83    \\
dense                    & \cmark & \cmark  &  \cmark  & \xmark &  \cmark     & 16.69$\pm$8.69     &  49.82$\pm$17.13    &   83.48$\pm$14.91   \\
w/o flow         & \cmark  & \cmark &  \cmark  & \cmark  & \xmark     & 13.00$\pm$6.99     &  58.36$\pm$17.47    &   86.41$\pm$ 16.40  \\       
\end{tabular}
\caption{Ablation study on StereoMIS. Metrics are reported as the mean $\pm$ std over all sequences.}
\label{tab:ablation}
\end{table}

{\bf Downstream application - 3D semantic segmentation:}
Our method achieves dense point tracking and coherent 3D scene reconstruction, facilitating downstream tasks like 3D semantic segmentation. We use the segmentation network defined in section~\ref{sec:seg} to infer semantic classes for each pixel, assigning them to Gaussians upon creation. Once assigned, semantics remain unchanged, enabling straightforward projection and propagation in 3D, as shown in Fig.~\ref{fig:semantic}. Visualizations for all scenes are available in the supplementary materials.
\begin{figure}[h]
    \centering
    \includegraphics[width=\textwidth]{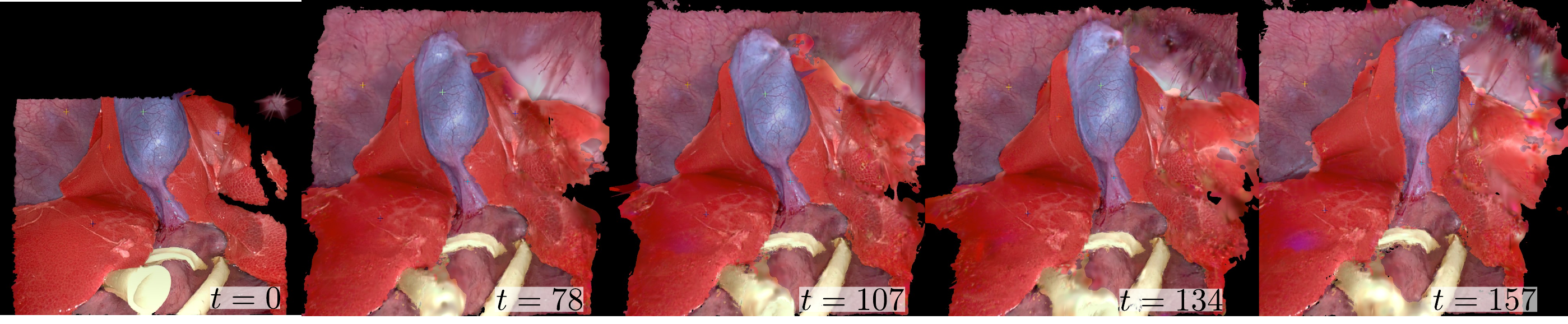}
    \caption{3D semantic segmentation as a downstream application. Semantic classes are overlayed: gall-bladder (purple), liver (red) and plastic tubes (yellow).}
    \label{fig:semantic}
\end{figure}

%% file: 4_conclusion.tex
\section{Conclusion}
We proposed a framework for online 3D scene reconstruction and dense tracking from stereo endoscopic video. To achieve this, we represent the scene as a collection of Gaussians that dynamically extend as the scene is explored and model tissue deformations through a sparse set of control points with physical priors. Through point tracking evaluation on the StereoMIS dataset, we validate the physical priors and demonstrate consistent online 3D reconstruction capability, outperforming state-of-the-art video tracking methods. We show the practicality of our framework on 3D semantic segmentation as a downstream application, highlighting its potential in surgical training, augmented reality overlays, and robotic assistance. Future efforts should focus on enhancing speed to achieve real-time processing, extending the method to long sequences and validating its robustness in real-world scenarios.